\documentclass{article}

    \usepackage[preprint]{neurips_2024}

\usepackage[utf8]{inputenc} 
\usepackage[T1]{fontenc} 
\usepackage[hyphens]{url}
\usepackage{hyperref}
\usepackage{url}
\usepackage{breakurl}
\usepackage{booktabs}       
\usepackage{amsfonts}       
\usepackage{nicefrac}       
\usepackage{microtype}      
\usepackage{xcolor}         
\usepackage{amsmath}
\usepackage{graphicx}
\usepackage{float}
\usepackage{array}
\usepackage{tabularx}

\title{A Comparative Study of Translation Bias and Accuracy in Multilingual Large Language Models for Cross-Language Claim Verification}

\author{%
Aryan Singhal \quad Veronica Shao \quad Gary Sun  \quad \textbf{Ryan Ding} \quad \textbf{Jonathan Lu} \quad \textbf{Kevin Zhu}
\\
Algoverse AI Research\\
\texttt{jonathan@algoverse.us, kevin@algoverse.us}
}

\begin{document}

\maketitle

\begin{abstract}
  % The abstract paragraph should be indented \nicefrac{1}{2}~inch (3~picas) on
  % both the left- and right-hand margins. Use 10~point type, with a vertical
  % spacing (leading) of 11~points.  The word \textbf{Abstract} must be centered,
  % bold, and in point size 12. Two line spaces precede the abstract. The abstract
  % must be limited to one paragraph.

The rise of digital misinformation has heightened interest in using multilingual Large Language Models (LLMs) for fact-checking. This study systematically evaluates translation bias and the effectiveness of LLMs for cross-lingual claim verification across 15 languages from five language families: Romance, Slavic, Turkic, Indo-Aryan, and Kartvelian. Using the XFACT dataset to assess their impact on accuracy and bias, we investigate two distinct translation methods: pre-translation and self-translation. We use mBERT’s performance on the English dataset as a baseline to compare language-specific accuracies. Our findings reveal that low-resource languages exhibit significantly lower accuracy in direct inference due to underrepresentation in the training data. Furthermore, larger models demonstrate superior performance in self-translation, improving translation accuracy and reducing bias. These results highlight the need for balanced multilingual training, especially in low-resource languages, to promote equitable access to reliable fact-checking tools and minimize the risk of spreading misinformation in different linguistic contexts.
\end{abstract}
\section{Introduction}
Multilingual Large Language Models (LLMs), such as GPT-4 and Llama 3.1, have shown remarkable capabilities in various languages and tasks \citep{ahuja2024megaversebenchmarkinglargelanguage}. Thus, there has been increasing interest in possible usages of LLMs for claim verification across languages \citep{Panchendrarajan_2024}.

However, recent studies have revealed significant disparities in their performance and bias in different languages \citep{xu2024surveymultilinguallargelanguage, huang2024surveylargelanguagemodels}. This variability is especially concerning given the importance of claim verification in combating misinformation \citep{sundriyal-etal-2023-chaos}. The performance discrepancies observed in LLMs often favor resource-rich languages like English, French, and German over resource-poor languages such as Kannada and Occitan \citep{robinson-etal-2023-chatgpt, bawden-yvon-2023-investigating, Quelle_2024}. These differences stem from variations in accuracy and translation quality between languages. Although LLMs demonstrate impressive average performance in a wide range of languages, \citet{li2024quantifyingmultilingualperformancelarge} highlights persistent gaps between high-resource and low-resource languages, emphasizing the need for more balanced data collection and training approaches.

Addressing misinformation for claim verification tasks is critical, as ineffective claim verification can spread false information between languages and vulnerable populations \citep{thorne-vlachos-2018-automated}. Although advances in LLMs, such as Meta's Llama 3.1 models \citep{dubey2024Llama3herdmodels}, have improved multilingual capabilities, reliance on external translation methods in some contexts---especially by users or systems that use third-party services such as Google Translate or that rely on the LLM in use and its multilingual capabilities---can still introduce biases. These biases can undermine the improvements made by LLMs and contribute to the spread of misinformation, particularly in resource-poor languages. Ensuring fair and accurate fact-checking in multiple languages is essential for equitable access to reliable information worldwide \citep{zhang2024needlanguagespecificfactcheckingmodels}.

This study evaluates pre-translation and self-translation methods across 15 languages, grouped into five language families---Romance, Slavic, Turkic, Indo-Aryan, and Kartvelian---spanning both high- and low-resource languages. We use mBERT's performance on the English dataset as a baseline to measure language-specific accuracy and the effectiveness of translation. The translation techniques are further explained in Section \ref{techniques} and are evaluated against the XFACT dataset by \citet{gupta2021xfactnewbenchmarkdataset}. Our analysis aims to inform the development of more balanced LLMs and guide future efforts in claim verification, helping to close the performance gap between high- and low-resource languages and creating more equitable language technologies.

\section{Related Works}
\subsection{\textbf{English and Multilingual Fact-Checking}}
The application of LLMs for fact-checking tasks has emerged as a promising area of research. \citet{Quelle_2024} demonstrated that the GPT-3.5 and GPT-4 models can achieve high accuracy in English fact-checking tasks when provided with adequate context. However, the challenge of extending these capabilities across multiple languages has driven research towards multilingual approaches. For example, \citet{huang-etal-2022-concrete} enhanced mBERT with cross-lingual retrieval techniques, improving fact-checking performance in the X-Fact dataset. \citet{hu2023largelanguagemodelsknow} further evaluated the factual knowledge of ten different LLMs in 27 languages, revealing insights into the multilingual capabilities of these models. Despite these advances, many studies have grouped non-English languages into a single category without detailed analysis, leaving a gap for users who wish or need to use other under-researched languages.

\subsection{\textbf{Bias in Multilingual Language Models}}
Wealthier countries often support more LLM research, leading to an uneven distribution of training data favoring their languages \citep{dong2024evaluatingmitigatinglinguisticdiscrimination, stanford2023nonenglishllms}. LLMs also exhibit political and informational biases, emphasizing claims spread by the media in wealthy countries over those in low-income countries. \citet{shafayat2024multifactassessingmultilingualllms} highlighted a significant bias toward Western-centric political information in the factual accuracy of LLMs across nine languages. Moreover, these models tend to produce more factual content in high-resource languages and longer responses in English. 

\section{Experimental Setup}
\begin{figure*}[!t]
    \centering
    \includegraphics[width=\textwidth]{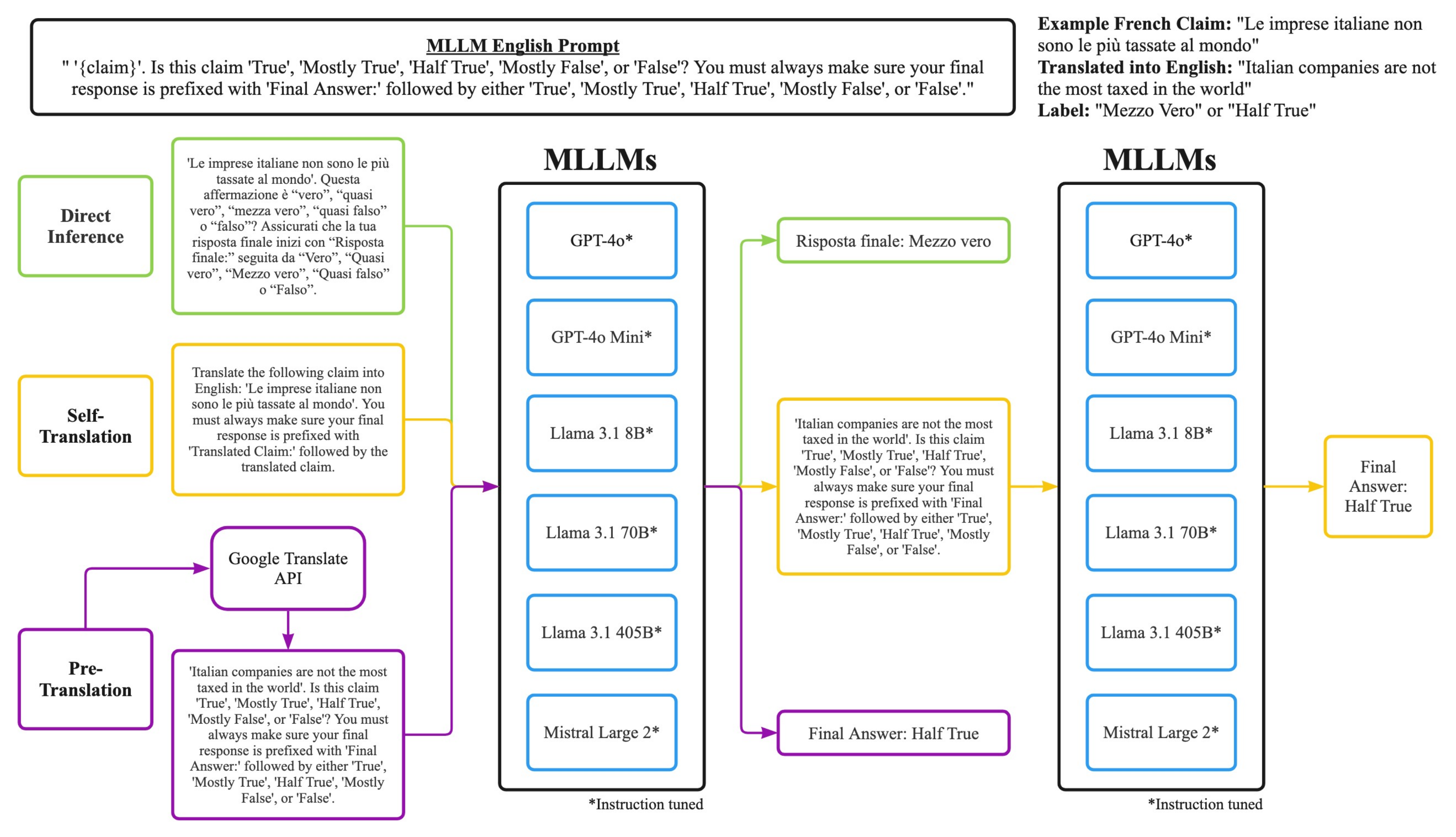}
    \caption{Flowchart illustrating the process for evaluating the claim verification performance of LLMs using Direct Inference, Self-Translation, and Pre-Translation.}
    \label{fig:flowchart}
\end{figure*}

\subsection{Datasets}
Our study uses the X-Fact dataset\footnote{X-Fact dataset under MIT License on GitHub (\url{https://github.com/utahnlp/x-fact})} developed by \citet{gupta2021xfactnewbenchmarkdataset} as the primary source of claims in selected language families. We systematically source 600 claims for each language family, ensuring a balanced representation of languages within each family and an equal distribution across the dataset's five veracity labels: "True", "Mostly True", "Half True", "Mostly False", and "False". The claims were selected to maintain an even distribution across both languages and veracity labels. This allowed for a diverse corpus encompassing both political and non-political topics. A detailed breakdown of the languages included in each family and the final dataset distribution is provided in Appendix \ref{sec:distribution}.

\subsection{Multilingual Language Models}
Each of the LLMs used in our experiments is instruction-tuned. We conduct our experiments on OpenAI's GPT-4o\footnote{\url{https://openai.com/index/hello-gpt-4o/}} and GPT-4o Mini\footnote{\url{https://openai.com/index/gpt-4o-mini-advancing-cost-efficient-intelligence/}} models, Mistral's Mistral Large 2\footnote{\url{https://mistral.ai/news/mistral-large-2407/}} model with 123B parameters, Meta's Llama 3.1 models with 8B, 70B, and 405B parameters \citep{dubey2024Llama3herdmodels}, and a fine-tuned version of Google's mBERT multilingual model, following the same training process used by \citet{gupta2021xfactnewbenchmarkdataset}. All of the models are pre-trained on multilingual corpora. For each model, we set the temperature to 0 for reproducibility. Each model automatically determined the default token length based on the number of tokens required to complete its output according to its respective context length.

\subsection{Evaluation}
For each experiment, we record the number of correct, incorrect, and inconclusive responses returned by the model. We express the accuracy score of the LLM as the percentage of correct answers.

\subsection{Translation Techniques}
\label{techniques}
We employ the following translation methods when evaluating each model's performance on a language family:

\noindent\textbf{Direct Inference} is completing a task in the native language of the prompt without performing any translations. This method is intended to measure the model's ability to understand and generate text in the target language without relying on cross-linguistic skills, thereby isolating its performance on monolingual tasks. Inconclusive outputs in this method occur when the model fails to provide a conclusive answer (e.g., "True," "Mostly True," etc.) as required by the prompt, though the risk of faulty translations is minimized since no external translations are involved.

\noindent\textbf{Self-Translate} \citet{etxaniz2023multilinguallanguagemodelsthink} involves an LLM performing a translation task itself without relying on external translation services. This technique allows the model to leverage its inherent multilingual capabilities, effectively using its own understanding of multiple languages to translate text autonomously. For consistent comparisons, we translate into English. This decision reflects the fact that most LLMs are trained predominantly on English data, making it a reasonable default for translation tasks, since translation into less well-trained languages is unlikely to yield better results due to the scarcity of high-quality training data in those languages \citep{dong2024evaluatingmitigatinglinguisticdiscrimination, stanford2023nonenglishllms}. The translation and claim verification steps are conducted in two separate chat sessions, ensuring that context is not preserved between them. This approach allows us to consistently assess the LLM's inherent translation ability, independent of any contextual memory. Inconclusive responses in this method occur if the model fails to properly translate the claim or does not follow the prompt’s instructions, resulting in incorrect or incomplete outputs.

\noindent\textbf{Pre-Translate} \citet{intrator2024breakinglanguagebarrierdirect} involves the use of third-party translation services external to the model, rather than relying on the model's own translation capabilities. Following the approach outlined by \citet{intrator2024breakinglanguagebarrierdirect}, we use the Google Translate API\footnote{\url{https://py-googletrans.readthedocs.io/en/latest/}} for this purpose. For consistent comparisons, we translate into English. In this method, inconclusive outputs can arise when there are inaccuracies in translation, which may lead the model to misinterpret the claim and provide an unclear or incorrect answer.

The process for evaluating claim verification performance using Direct Inference, Self-Translation, and Pre-Translation is outlined in the flowchart shown in Figure \ref{fig:flowchart}.

\subsection{Translation Bias}
We assess translation bias using the COMETKIWI model from \citet{rei-etal-2022-cometkiwi}, which allows for the evaluation of machine translations without requiring reference translations. A reference translation is a pre-existing human translation of a source text that serves as a benchmark for evaluating the accuracy and quality of a machine translation.

The Translation Bias (TB) quantifies the overall quality of machine translations by leveraging the scores from the COMETKIWI model. Given a set of $M$ COMET scores, $\text{scores} = \{s_1, s_2, \ldots, s_M\}$, the Translation Bias is calculated as:
\[
\text{TB} = 1 - \frac{1}{M} \sum_{j=1}^{M} s_j
\]

\section{Results and Discussion}
\subsection{Language-Specific Trends}
\subsubsection{High- and Low-Resource Languages}
Direct inference demonstrated significantly higher accuracy in the Romance, Slavic, and Turkic language families compared to other translation techniques. These families generally consist of high- or moderately high-resource languages, which have abundant data and represent a larger portion of the training data for the models. In contrast, for the Kartvelian and Indo-Aryan language families---mostly low-resource languages---the performance of direct inference was consistently equal to or worse than other translation methods. This suggests that direct inference may be less effective for low-resource languages due to limited training data, resulting in poorer model understanding and higher error rates.

\subsubsection{Performance of English}
Despite being the most represented language in the training data, English was sometimes outperformed by other languages, possibly because the English claims in the evaluation dataset are more niche and complex---often including a higher proportion of political claims---which may lower accuracy as models struggle with more intricate statements. For instance, Llama 3.1 405B showed higher accuracy for Slavic languages (36.00\%) compared to English (33.50\%), even though English is typically better resourced.

\subsection{Translation Techniques}
While self-translation and pre-translation techniques generally yielded lower accuracy compared to direct inference, they reduced the number of inconclusive results by enhancing LLM comprehension and likely reducing misinterpretations, particularly for complex or nuanced claims. Nonetheless, the accuracy of both translation methods remained lower than that of direct inference.

\subsubsection{Self-Translation vs. Pre-Translation}
Self-translation performs slightly better than pre-translation which we believe is attributed to the model maintaining internal consistency between generating and verifying translations. When the LLM handles both tasks, its linguistic patterns are more likely to align, reducing interpretation errors. Pre-translation, however, relies on external services that can introduce inconsistencies, leading to more misinterpretations during the verification phase. As a result, pre-translation produced more inconclusive outputs and had lower accuracy than self-translation.

\subsection{Model Scale}
\begin{figure}[!t]
    \centering
    \includegraphics[width=\columnwidth]{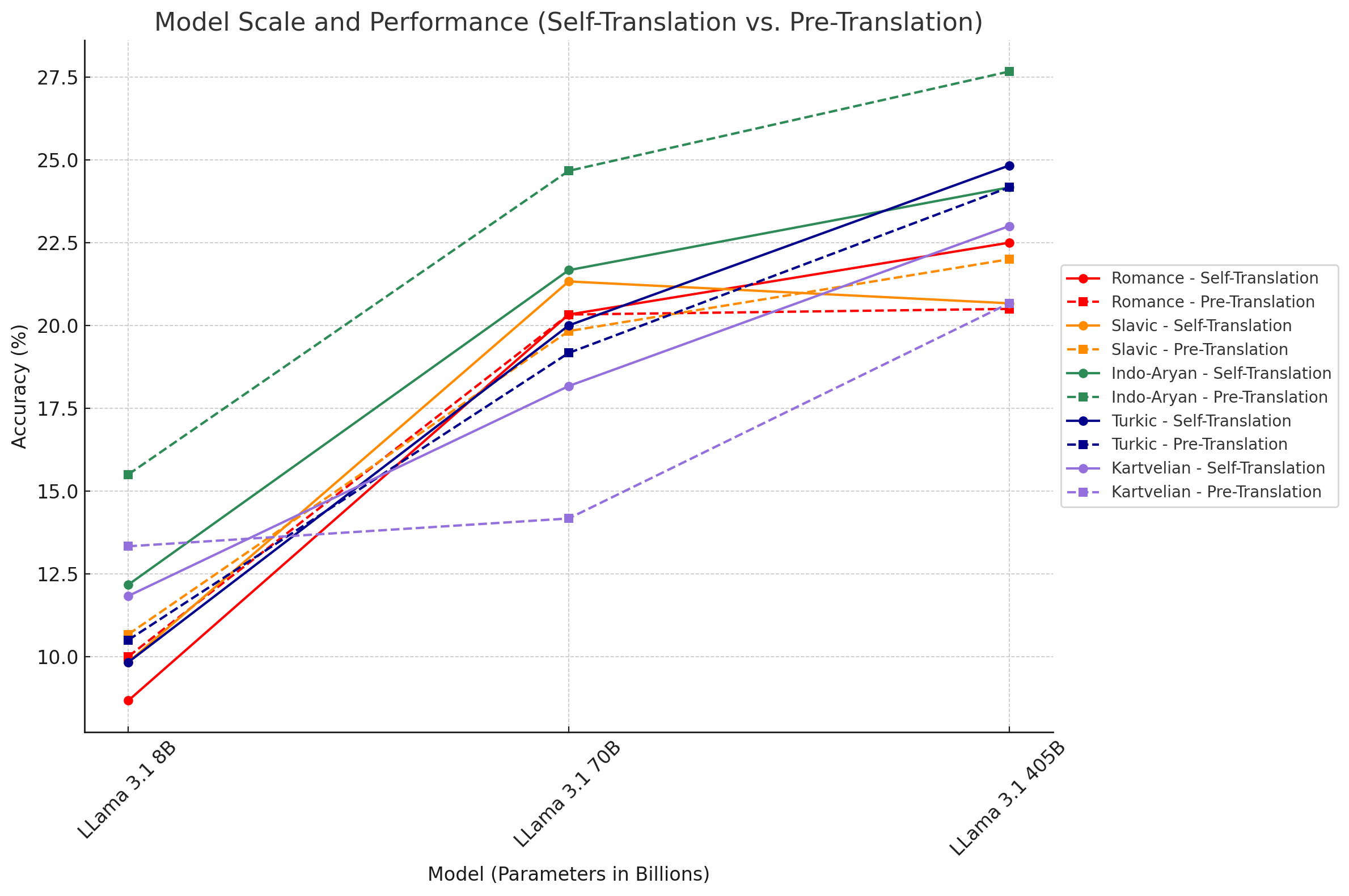}
    \caption{Accuracy performance of Llama 3.1 models across different language families using Self-Translation and Pre-Translation techniques.}
    \label{fig:scale_perf}
\end{figure}

Looking at Figure~\ref{fig:scale_perf}, smaller models like Llama 3.1 8B perform poorly for both self-translation and pre-translation across all language families, with self-translation slightly outperforming pre-translation. However, as model size increased, the accuracy of self-translation improved significantly. For instance, Llama 3.1 405B demonstrated improved performance across Romance, Slavic, Turkic, and Indo-Aryan languages, surpassing pre-translation in all cases.

Interestingly, although self-translation performed better with larger models, the translation bias scores remained relatively stable, suggesting that increased model size improves accuracy but not fairness across languages. For example, Llama 3.1 405B maintained similar bias scores to smaller models like Llama 3.1 8B, indicating that the increased size of the model improves accuracy but not fairness in translation. A detailed breakdown of the translation bias scores for each method, model, and language family is provided in Appendix \ref{sec:translation_bias}.

\section{Conclusion}
This study examines the translation bias and accuracy of multilingual Large Language Models (LLMs) in cross-language claim verification tasks across five language families. Our findings demonstrate that direct inference performs better in high-resource languages, while self-translation and pre-translation techniques handle low-resource languages more effectively, though with reduced accuracy. Furthermore, as model size increases, the accuracy of self-translation improves, yet translation bias remains consistent across all models, showing that larger models do not necessarily ensure fairness across languages. These results highlight the persistent challenges in achieving equitable multilingual capabilities in LLMs. By identifying specific areas where translation biases occur, we lay the groundwork for developing more balanced and fair language technologies.

\section*{Limitations}
Our study of language and translation biases in LLMs for cross-lingual claim verification has several limitations. We used the 2021 X-Fact dataset, which may not reflect the most recent language trends or advancements in model capabilities as of 2024. Additionally, the LLMs tested may have been trained on datasets overlapping with X-Fact, potentially inflating performance metrics. While we focused on 15 languages from diverse families, this selection might not fully represent the linguistic diversity needed to capture trends in low-resource languages. Our evaluation was limited to translations from non-English languages into English, and while examining other language pairs might provide valuable insights, it is unlikely that these pairings would outperform English due to the prevalent training bias toward English data in most LLMs. We used a lighter, older version of the COMETKIWI model to assess translation bias due to computational limitations, which may affect the robustness of our bias measurements. Moreover, we did not compare baseline models with instruction-tuned versions, which could have reduced inconclusive translations and offered further insights into model performance. We also did not incorporate reference translations or employ evidence retrieval, which could have provided a more holistic evaluation of translation quality. Future work should expand to include more recent datasets, evaluate other language pairs, and refine the methods to enhance bias detection and accuracy.

\section*{Ethics Statement}
This study investigates translation bias in multilingual Large Language Models (LLMs), focusing on disparities across high- and low-resource languages. Our findings highlight that these biases disproportionately affect low-resource languages, potentially leading to misinformation propagation in underrepresented linguistic communities. We acknowledge the potential ethical risks associated with the reliance on LLMs for cross-language claim verification, particularly the unequal access to accurate information. Future work should focus on more balanced model training to mitigate these risks, ensuring fairer outcomes for all language speakers. Additionally, we emphasize the need for collaboration with native speakers and ethical oversight in model development to ensure inclusivity in global language technologies.

% References section
\bibliographystyle{unsrtnat}
\bibliography{neurips_2024}

\appendix

\section{Appendix}

% Optionally include supplemental material (complete proofs, additional experiments and plots) in appendix.
% All such materials \textbf{SHOULD be included in the main submission.}

\subsection{Dataset Distribution}
\label{sec:distribution}

\begin{table}[H]
\centering
\caption{Distribution of Romance language family claims.}
\resizebox{\columnwidth}{!}{
\begin{tabular}{|l|c|c|c|c|c|}
\hline
\textbf{Language} & \textbf{False} & \textbf{Half True} & \textbf{Mostly False} & \textbf{True} & \textbf{Total Claims} \\ \hline
French (fr)       & 30   & 30   & 0    & 21  & 109  \\ \hline
Italian (it)      & 34   & 36   & 0    & 37  & 139  \\ \hline
Spanish (es)      & 34   & 34   & 0    & 37  & 136  \\ \hline
Portuguese (pt)   & 80   & 43   & 1    & 50  & 216  \\ \hline
\end{tabular}
}
\end{table}

\begin{table}[H]
\centering
\caption{Distribution of Slavic language family claims.}
\resizebox{\columnwidth}{!}{%
\begin{tabular}{|l|c|c|c|c|c|}
\hline
\textbf{Language} & \textbf{False} & \textbf{Half True} & \textbf{Mostly False} & \textbf{True} & \textbf{Total Claims} \\ \hline
Serbian (sr)      & 70   & 42   & 44   & 45  & 234  \\ \hline
Russian (ru)      & 50   & 51   & 1    & 42  & 153  \\ \hline
Polish (pl)       & 64   & 50   & 0    & 99  & 213  \\ \hline
\end{tabular}%
}
\end{table}

\begin{table}[H]
\centering
\caption{Distribution of Turkic language family claims.}
\resizebox{\columnwidth}{!}{%
\begin{tabular}{|l|c|c|c|c|c|}
\hline
\textbf{Language} & \textbf{False} & \textbf{Half True} & \textbf{Mostly False} & \textbf{True} & \textbf{Total Claims} \\ \hline
Turkish (tr)      & 60   & 63   & 82   & 96  & 407  \\ \hline
Azerbaijani (az)  & 60   & 57   & 38   & 24  & 193  \\ \hline
\end{tabular}%
}
\end{table}

\begin{table}[H]
\centering
\caption{Distribution of Indo-Aryan language family claims.}
\resizebox{\columnwidth}{!}{%
\begin{tabular}{|l|c|c|c|c|c|}
\hline
\textbf{Language} & \textbf{False} & \textbf{Half True} & \textbf{Mostly False} & \textbf{True} & \textbf{Total Claims} \\ \hline
Bengali (bn)      & 36   & 35   & 91   & 1   & 163  \\ \hline
Hindi (hi)        & 89   & 57   & 118  & 0   & 264  \\ \hline
Marathi (mr)      & 26   & 26   & 0    & 0   & 52   \\ \hline
Punjabi (pa)      & 25   & 40   & 0    & 0   & 65   \\ \hline
Gujarati (gu)     & 27   & 29   & 0    & 0   & 56   \\ \hline
\end{tabular}%
}
\end{table}

\begin{table}[H]
\centering
\caption{Distribution of Kartvelian language family claims.}
\resizebox{\columnwidth}{!}{%
\begin{tabular}{|l|c|c|c|c|c|}
\hline
\textbf{Language} & \textbf{False} & \textbf{Half True} & \textbf{Mostly False} & \textbf{True} & \textbf{Total Claims} \\ \hline
Georgian (ka)     & 120  & 120  & 120  & 120 & 600  \\ \hline
\end{tabular}%
}
\end{table}

\subsection{Model Performance Across Language Families}
\label{sec:model_performance}
\begin{table}[H]
\centering
\caption{Performance distribution of LLMs using direct inference on English claims.}
\resizebox{\columnwidth}{!}{
\begin{tabular}{|l|c|c|c|c|}
\hline
\textbf{Model}      & \textbf{Total Correct} & \textbf{Total Incorrect} & \textbf{Total Inconclusive} & \textbf{Accuracy} \\ \hline
GPT-4o              & 215                    & 377                      & 8                           & \textbf{35.83\%}  \\ \hline
GPT-4o Mini         & 185                    & 413                      & 2                           & 30.83\%          \\ \hline
Mistral Large 2     & 183                    & 403                      & 14                          & 30.50\%          \\ \hline
Llama 3.1 8B        & 95                     & 278                      & 227                         & 15.83\%          \\ \hline
Llama 3.1 70B       & 166                    & 353                      & 81                          & 27.67\%          \\ \hline
Llama 3.1 405B      & 201                    & 386                      & 13                          & 33.50\%          \\ \hline
mBERT               & 95                     & 340                      & 165                         & 15.83\%          \\ \hline
\end{tabular}
}
\end{table}
\begin{table}[H]
\centering
\caption{Performance distribution of LLMs using direct inference, self-translate, and pre-translate on Romance claims.}
\resizebox{\columnwidth}{!}{
\begin{tabular}{|l|c|c|c|c|c|}
\hline
\textbf{Model}      & \textbf{Technique}      & \textbf{Total Correct} & \textbf{Total Incorrect} & \textbf{Total Inconclusive} & \textbf{Accuracy} \\ \hline
GPT-4o              & Direct Inference        & 185                    & 381                      & 34                          & 30.83\%          \\ \hline
GPT-4o              & Self-Translation        & 174                    & 396                      & 30                          & 29.00\%          \\ \hline
GPT-4o              & Pre-Translation         & 150                    & 413                      & 37                          & 25.00\%          \\ \hline
GPT-4o Mini         & Direct Inference        & 197                    & 388                      & 15                          & \textbf{32.83\%} \\ \hline
GPT-4o Mini         & Self-Translation        & 165                    & 434                      & 1                           & \textbf{27.50\%} \\ \hline
GPT-4o Mini         & Pre-Translation         & 154                    & 445                      & 1                           & 25.67\%          \\ \hline
Mistral Large 2     & Direct Inference        & 155                    & 405                      & 40                          & 25.83\%          \\ \hline
Mistral Large 2     & Self-Translation        & 123                    & 422                      & 55                          & 20.50\%          \\ \hline
Mistral Large 2     & Pre-Translation         & 97                     & 386                      & 117                         & 16.17\%          \\ \hline
Llama 3.1 8B        & Direct Inference        & 126                    & 389                      & 85                          & 21.00\%          \\ \hline
Llama 3.1 8B        & Self-Translation        & 52                     & 236                      & 312                         & 8.67\%           \\ \hline
Llama 3.1 8B        & Pre-Translation         & 60                     & 296                      & 244                         & 10.00\%          \\ \hline
Llama 3.1 70B       & Direct Inference        & 172                    & 398                      & 30                          & 28.67\%          \\ \hline
Llama 3.1 70B       & Self-Translation        & 122                    & 303                      & 175                         & 20.33\%          \\ \hline
Llama 3.1 70B       & Pre-Translation         & 122                    & 301                      & 177                         & 20.33\%          \\ \hline
Llama 3.1 405B      & Direct Inference        & 191                    & 404                      & 5                           & \textbf{31.83\%} \\ \hline
Llama 3.1 405B      & Self-Translation        & 135                    & 422                      & 43                          & 22.50\%          \\ \hline
Llama 3.1 405B      & Pre-Translation         & 123                    & 427                      & 50                          & 20.50\%          \\ \hline
mBERT               & Direct Inference        & 166                    & 255                      & 179                         & 27.67\%          \\ \hline
mBERT               & Pre-Translation         & 106                    & 444                      & 50                          & 17.67\%          \\ \hline
\end{tabular}
}
\end{table}

\begin{table}[H]
\centering
\caption{Performance distribution of LLMs using direct inference, self-translate, and pre-translate on Slavic claims.}
\resizebox{\columnwidth}{!}{
\begin{tabular}{|l|c|c|c|c|c|}
\hline
\textbf{Model}      & \textbf{Technique}      & \textbf{Total Correct} & \textbf{Total Incorrect} & \textbf{Total Inconclusive} & \textbf{Accuracy} \\ \hline
GPT-4o              & Direct Inference        & 199                    & 315                      & 86                          & 33.17\%          \\ \hline
GPT-4o              & Self-Translation        & 195                    & 384                      & 21                          & \textbf{32.50\%} \\ \hline
GPT-4o              & Pre-Translation         & 161                    & 404                      & 35                          & 26.83\%          \\ \hline
GPT-4o Mini         & Direct Inference        & 206                    & 334                      & 60                          & \textbf{34.33\%} \\ \hline
GPT-4o Mini         & Self-Translation        & 135                    & 465                      & 0                           & 22.50\%          \\ \hline
GPT-4o Mini         & Pre-Translation         & 139                    & 461                      & 0                           & 23.17\%          \\ \hline
Mistral Large 2     & Direct Inference        & 177                    & 298                      & 125                         & 29.50\%          \\ \hline
Mistral Large 2     & Self-Translation        & 123                    & 439                      & 38                          & 20.50\%          \\ \hline
Mistral Large 2     & Pre-Translation         & 102                    & 423                      & 75                          & 17.00\%          \\ \hline
Llama 3.1 8B        & Direct Inference        & 121                    & 250                      & 229                         & 20.17\%          \\ \hline
Llama 3.1 8B        & Self-Translation        & 59                     & 253                      & 288                         & 9.83\%           \\ \hline
Llama 3.1 8B        & Pre-Translation         & 64                     & 280                      & 256                         & 10.67\%          \\ \hline
Llama 3.1 70B       & Direct Inference        & 177                    & 290                      & 133                         & 29.50\%          \\ \hline
Llama 3.1 70B       & Self-Translation        & 128                    & 357                      & 115                         & 21.33\%          \\ \hline
Llama 3.1 70B       & Pre-Translation         & 119                    & 388                      & 93                          & 19.83\%          \\ \hline
Llama 3.1 405B      & Direct Inference        & 216                    & 353                      & 31                          & 36.00\%          \\ \hline
Llama 3.1 405B      & Self-Translation        & 124                    & 468                      & 8                           & 20.67\%          \\ \hline
Llama 3.1 405B      & Pre-Translation         & 132                    & 454                      & 14                          & \textbf{22.00\%} \\ \hline
mBERT               & Direct Inference        & 79                     & 251                      & 270                         & 13.17\%          \\ \hline
mBERT               & Pre-Translation         & 130                    & 411                      & 59                          & 21.67\%          \\ \hline
\end{tabular}
}
\end{table}

\begin{table}[H]
\centering
\caption{Performance distribution of LLMs using direct inference, self-translate, and pre-translate on Indo-Aryan claims.}
\resizebox{\columnwidth}{!}{
\begin{tabular}{|l|c|c|c|c|c|}
\hline
\textbf{Model}      & \textbf{Technique}      & \textbf{Total Correct} & \textbf{Total Incorrect} & \textbf{Total Inconclusive} & \textbf{Accuracy} \\ \hline
GPT-4o              & Direct Inference        & 150                    & 425                      & 25                          & 25.00\%          \\ \hline
GPT-4o              & Self-Translation        & 180                    & 360                      & 60                          & \textbf{30.00\%} \\ \hline
GPT-4o              & Pre-Translation         & 157                    & 346                      & 97                          & 26.17\%          \\ \hline
GPT-4o Mini         & Direct Inference        & 190                    & 431                      & 0                           & \textbf{28.17\%} \\ \hline
GPT-4o Mini         & Self-Translation        & 144                    & 434                      & 0                           & 27.67\%          \\ \hline
GPT-4o Mini         & Pre-Translation         & 171                    & 429                      & 0                           & \textbf{28.50\%} \\ \hline
Mistral Large 2     & Direct Inference        & 85                     & 281                      & 234                         & 14.17\%          \\ \hline
Mistral Large 2     & Self-Translation        & 173                    & 364                      & 63                          & 28.83\%          \\ \hline
Mistral Large 2     & Pre-Translation         & 146                    & 300                      & 154                         & 24.33\%          \\ \hline
Llama 3.1 8B        & Direct Inference        & 95                     & 278                      & 227                         & 15.83\%          \\ \hline
Llama 3.1 8B        & Self-Translation        & 73                     & 192                      & 335                         & 12.17\%          \\ \hline
Llama 3.1 8B        & Pre-Translation         & 93                     & 222                      & 285                         & 15.50\%          \\ \hline
Llama 3.1 70B       & Direct Inference        & 127                    & 426                      & 47                          & 21.17\%          \\ \hline
Llama 3.1 70B       & Self-Translation        & 130                    & 344                      & 126                         & 21.67\%          \\ \hline
Llama 3.1 70B       & Pre-Translation         & 148                    & 321                      & 131                         & 24.67\%          \\ \hline
Llama 3.1 405B      & Direct Inference        & 166                    & 358                      & 76                          & 27.67\%          \\ \hline
Llama 3.1 405B      & Self-Translation        & 143                    & 379                      & 76                          & 24.17\%          \\ \hline
Llama 3.1 405B      & Pre-Translation         & 166                    & 358                      & 76                          & 27.67\%          \\ \hline
mBERT               & Direct Inference        & 81                     & 279                      & 240                         & 13.50\%          \\ \hline
mBERT               & Pre-Translation         & 83                     & 281                      & 216                         & 17.17\%          \\ \hline
\end{tabular}
}
\end{table}
\begin{table}[H]
\centering
\caption{Performance distribution of LLMs using direct inference, self-translate, and pre-translate on Turkic claims.}
\resizebox{\columnwidth}{!}{
\begin{tabular}{|l|c|c|c|c|c|}
\hline
\textbf{Model}      & \textbf{Technique}      & \textbf{Total Correct} & \textbf{Total Incorrect} & \textbf{Total Inconclusive} & \textbf{Accuracy} \\ \hline
GPT-4o              & Direct Inference        & 159                    & 437                      & 4                           & 26.50\%          \\ \hline
GPT-4o              & Self-Translation        & 150                    & 416                      & 34                          & 25.00\%          \\ \hline
GPT-4o              & Pre-Translation         & 141                    & 427                      & 32                          & 23.50\%          \\ \hline
GPT-4o Mini         & Direct Inference        & 130                    & 469                      & 1                           & 21.67\%          \\ \hline
GPT-4o Mini         & Self-Translation        & 147                    & 452                      & 1                           & \textbf{24.50\%} \\ \hline
GPT-4o Mini         & Pre-Translation         & 135                    & 462                      & 3                           & 22.50\%          \\ \hline
Mistral Large 2     & Direct Inference        & 129                    & 469                      & 2                           & 21.50\%          \\ \hline
Mistral Large 2     & Self-Translation        & 123                    & 418                      & 59                          & 20.50\%          \\ \hline
Mistral Large 2     & Pre-Translation         & 111                    & 396                      & 93                          & 18.50\%          \\ \hline
Llama 3.1 8B        & Direct Inference        & 106                    & 454                      & 40                          & 17.67\%          \\ \hline
Llama 3.1 8B        & Self-Translation        & 59                     & 247                      & 294                         & 9.83\%           \\ \hline
Llama 3.1 8B        & Pre-Translation         & 63                     & 307                      & 230                         & 10.50\%          \\ \hline
Llama 3.1 70B       & Direct Inference        & 131                    & 443                      & 26                          & 21.83\%          \\ \hline
Llama 3.1 70B       & Self-Translation        & 120                    & 359                      & 121                         & 20.00\%          \\ \hline
Llama 3.1 70B       & Pre-Translation         & 115                    & 379                      & 106                         & 19.17\%          \\ \hline
Llama 3.1 405B      & Direct Inference        & 154                    & 445                      & 1                           & \textbf{25.67\%} \\ \hline
Llama 3.1 405B      & Self-Translation        & 149                    & 432                      & 19                          & 24.83\%          \\ \hline
Llama 3.1 405B      & Pre-Translation         & 145                    & 439                      & 16                          & 24.17\%          \\ \hline
mBERT               & Direct Inference        & 98                     & 331                      & 171                         & 16.33\%          \\ \hline
mBERT               & Pre-Translation         & 109                    & 478                      & 13                          & \textbf{18.17\%} \\ \hline
\end{tabular}
}
\end{table}
\begin{table}[H]
\centering
\caption{Performance distribution of LLMs using direct inference, self-translate, and pre-translate on Kartvelian claims.}
\resizebox{\columnwidth}{!}{
\begin{tabular}{|l|c|c|c|c|c|}
\hline
\textbf{Model}      & \textbf{Technique}      & \textbf{Total Correct} & \textbf{Total Incorrect} & \textbf{Total Inconclusive} & \textbf{Accuracy} \\ \hline
GPT-4o              & Direct Inference        & 28                     & 503                      & 69                          & 4.67\%           \\ \hline
GPT-4o              & Self-Translation        & 131                    & 423                      & 46                          & 21.83\%\\ \hline
GPT-4o              & Pre-Translation         & 127                    & 442                      & 31                          & 21.17\%          \\ \hline
GPT-4o Mini         & Direct Inference        & 38                     & 559                      & 3                           & 6.33\%           \\ \hline
GPT-4o Mini         & Self-Translation        & 138                    & 459                      & 3                           & \textbf{23.00\%} \\ \hline
GPT-4o Mini         & Pre-Translation         & 132                    & 465                      & 3                           & \textbf{22.00\%} \\ \hline
Mistral Large 2     & Direct Inference        & 42                     & 303                      & 255                         & 7.00\%           \\ \hline
Mistral Large 2     & Self-Translation        & 118                    & 404                      & 78                          & 19.67\%          \\ \hline
Mistral Large 2     & Pre-Translation         & 107                    & 386                      & 107                         & 17.83\%          \\ \hline
Llama 3.1 8B        & Direct Inference        & 29                     & 135                      & 436                         & 4.83\%           \\ \hline
Llama 3.1 8B        & Self-Translation        & 71                     & 236                      & 293                         & 11.83\%          \\ \hline
Llama 3.1 8B        & Pre-Translation         & 80                     & 267                      & 253                         & 13.33\%          \\ \hline
Llama 3.1 70B       & Direct Inference        & 55                     & 511                      & 34                          & 9.17\%           \\ \hline
Llama 3.1 70B       & Self-Translation        & 109                    & 336                      & 155                         & 18.17\%          \\ \hline
Llama 3.1 70B       & Pre-Translation         & 85                     & 313                      & 202                         & 14.17\%          \\ \hline
Llama 3.1 405B      & Direct Inference        & 0                      & 598                      & 0                           & 0.00\%           \\ \hline
Llama 3.1 405B      & Self-Translation        & 138                    & 435                      & 27                          & \textbf{23.00\%} \\ \hline
Llama 3.1 405B      & Pre-Translation         & 124                    & 439                      & 37                          & 20.67\%          \\ \hline
mBERT               & Direct Inference        & 133                    & 463                      & 4                           & \textbf{22.17\%}\\ \hline
mBERT               & Pre-Translation         & 99                     & 398                      & 103                         & 16.50\%          \\ \hline
\end{tabular}
}
\end{table}

\subsection{Translation Bias Scores}
\label{sec:translation_bias}
\begin{table}[H]
\centering
\caption{Translation bias scores across Romance, Slavic, Turkic, Indo-Aryan, and Kartvelian language families using the pre-translation technique (Google Translate API).}
{%
\begin{tabular}{|l|c|}
\hline
\textbf{Language Family}& \textbf{Translation Bias Score} \\ \hline
Romance     & 0.33  \\ \hline
Slavic      & 0.35  \\ \hline
Turkic      & 0.11  \\ \hline
Indo-Aryan  & 0.22  \\ \hline
Kartvelian  & 0.22  \\ \hline
\end{tabular}%
}
\end{table}
\begin{table}[H]
\centering
\caption{Translation bias scores for LLMs across Romance, Slavic, Turkic, Indo-Aryan, and Kartvelian language families using the self-translation technique.}
\resizebox{\columnwidth}{!}{%
\begin{tabular}{|l|l|c|}
\hline
\textbf{Model}     & \textbf{Language Family}  & \textbf{Translation Bias Score} \\ \hline
GPT-4o         & Romance     & 0.16  \\ \hline
GPT-4o         & Slavic      & 0.16  \\ \hline
GPT-4o         & Turkic      & 0.17  \\ \hline
GPT-4o         & Indo-Aryan  & 0.16  \\ \hline
GPT-4o         & Kartvelian  & 0.16  \\ \hline
GPT-4o Mini    & Romance     & 0.16  \\ \hline
GPT-4o Mini    & Slavic      & 0.16  \\ \hline
GPT-4o Mini    & Turkic      & 0.17  \\ \hline
GPT-4o Mini    & Indo-Aryan  & 0.17  \\ \hline
GPT-4o Mini    & Kartvelian  & 0.16  \\ \hline
Mistral Large 2 & Romance    & 0.16  \\ \hline
Mistral Large 2 & Slavic     & 0.16  \\ \hline
Mistral Large 2 & Turkic     & 0.17  \\ \hline
Mistral Large 2 & Indo-Aryan & 0.17  \\ \hline
Mistral Large 2 & Kartvelian & 0.17  \\ \hline
Llama 3.1 8B   & Romance     & 0.18  \\ \hline
Llama 3.1 8B   & Slavic      & 0.19  \\ \hline
Llama 3.1 8B   & Turkic      & 0.20  \\ \hline
Llama 3.1 8B   & Indo-Aryan  & 0.19  \\ \hline
Llama 3.1 8B   & Kartvelian  & 0.21  \\ \hline
Llama 3.1 70B  & Romance     & 0.16  \\ \hline
Llama 3.1 70B  & Slavic      & 0.17  \\ \hline
Llama 3.1 70B  & Turkic      & 0.18  \\ \hline
Llama 3.1 70B  & Indo-Aryan  & 0.18  \\ \hline
Llama 3.1 70B  & Kartvelian  & 0.19  \\ \hline
Llama 3.1 405B & Romance     & 0.16  \\ \hline
Llama 3.1 405B & Slavic      & 0.16  \\ \hline
Llama 3.1 405B & Turkic      & 0.17  \\ \hline
Llama 3.1 405B & Indo-Aryan  & 0.16  \\ \hline
Llama 3.1 405B & Kartvelian  & 0.16  \\ \hline
\end{tabular}%
}
\end{table}

\subsection{Code Repository}
The code used in our experiments and for generating the results presented in this paper can be accessed at the following GitHub repository:\\
\url{https://github.com/3x-dev/Comparative-Study-of-Bias-and-Accuracy-in-Multilingual-LLMs-for-Cross-Language-Claim-Verification}

\subsection{Compute Resources}
The experiments were conducted using a combination of MacBook Pros and a dedicated GPU cluster for pre-training the mBERT model. Below are the general specifications for each setup:

\textbf{GPU Resources:} The mBERT pre-training was performed on a GPU cluster equipped with NVIDIA A100 Tensor Core GPUs (40 GB VRAM) for high performance training. Inference and other experiments performed on MacBook Pros did not use GPUs because MacBook Pros do not have discrete GPUs suitable for machine learning tasks. \\

\textbf{CPU Resources:} Experiments run on MacBook Pros used Apple's \textbf{M1 Pro} or \textbf{M1 Max} processors (8- to 10-core CPUs), and some collaborators used \textbf{Intel Core i9} processors (8-core) in older MacBook Pro models. These CPU configurations were sufficient for smaller experiments and model inference tasks. \\

\textbf{Memory:} MacBook Pro memory capacity ranged from \textbf{16GB to 64GB of unified memory} on Apple Silicon (M1) models to \textbf{32GB of DDR4 RAM} on Intel-based MacBook Pros. These configurations were sufficient for model inference, but could limit performance with larger models and datasets. \\

\textbf{Storage:} Experiments conducted on MacBook Pros used \textbf{SSD storage ranging from 512GB to 2TB}. Local storage was used to manage smaller datasets and model checkpoints. For larger datasets and models, external storage or cloud services were used to mitigate local storage limitations. \\

\textbf{pre-training and Inference Times:} 
\begin{itemize}
    \item \textbf{pre-training:} pre-training mBERT on the GPU cluster with NVIDIA A100 GPUs took approximately \textbf{12 hours} using 4 GPUs in parallel. This was essential to ensure the mBERT model was fine-tuned for multilingual tasks.
    \item \textbf{Inference:} Inference on the MacBook Pros varied depending on model size. For smaller models like GPT-4 Mini, inference times ranged between \textbf{3 to 5 hours} per language family. However, larger models like Llama 3.1 405B were run in a distributed fashion, with inference times extending to \textbf{8 to 10 hours} due to limited hardware.
\end{itemize}

\textbf{Total Computing Time:} The total computation time for all experiments, including pre-training, tuning, and inference, was approximately \textbf{150 GPU hours} on the cluster for pre-training and \textbf{100 CPU hours} on MacBook Pros for inference and evaluation. \\

\textbf{Considerations for Reproducibility:} Replicating these results on similar hardware, such as MacBook Pros with M1/M2 chips or Intel processors, should result in longer computation times, especially for larger models. For pre-training or large-scale fine-tuning, access to a GPU cluster or cloud-based GPU services is recommended.
%%%%%%%%%%%%%%%%%%%%%%%%%%%%%%%%%%%%%%%%%%%%%%%%%%%%%%%%%%%%

\newpage
\section*{NeurIPS Paper Checklist}

\begin{enumerate}

\item {\bf Claims}
    \item[] Question: Do the main claims made in the abstract and introduction accurately reflect the paper's contributions and scope?
    \item[] Answer: \answerYes{} % Replace by \answerYes{}, \answerNo{}, or \answerNA{}.
    \item[] Justification: The abstract and introduction clearly state the focus on evaluating language and translation biases in LLMs for cross-lingual claim verification. The claims align with the experiments and analysis conducted, and the scope of the study is appropriately represented. See Abstract and Section 1.

\item {\bf Limitations}
    \item[] Question: Does the paper discuss the limitations of the work performed by the authors?
    \item[] Answer: \answerYes{}{} % Replace by \answerYes{}, \answerNo{}, or \answerNA{}.
    \item[] Justification: The paper contains a dedicated limitations section, which outlines key constraints such as dataset recency, reliance on English-centric training data, the exclusion of non-English language pairs, and the use of an older COMETKIWI model for bias evaluation. See the Limitations section.

\item {\bf Theory Assumptions and Proofs}
    \item[] Question: For each theoretical result, does the paper provide the full set of assumptions and a complete (and correct) proof?
    \item[] Answer: \answerNA{} % Replace by \answerYes{}, \answerNo{}, or \answerNA{}.
    \item[] Justification: The paper focuses on empirical results and does not present theoretical results or proofs.

    \item {\bf Experimental Result Reproducibility}
    \item[] Question: Does the paper fully disclose all the information needed to reproduce the main experimental results of the paper to the extent that it affects the main claims and/or conclusions of the paper (regardless of whether the code and data are provided or not)?
    \item[] Answer: \answerYes{} % Replace by \answerYes{}, \answerNo{}, or \answerNA{}.
    \item[] Justification: The experimental setup is well documented with details on datasets, models, evaluation techniques, and hyperparameters. The methods section (Section 3) provides enough information to reproduce the experiments.

\item {\bf Open access to data and code}
    \item[] Question: Does the paper provide open access to the data and code, with sufficient instructions to faithfully reproduce the main experimental results, as described in supplemental material?
    \item[] Answer: \answerYes{} % Replace by \answerYes{}, \answerNo{}, or \answerNA{}.
    \item[] Justification: The paper links to a GitHub repository containing the code and scripts used in the experiments. See Appendix A.4.

\item {\bf Experimental Setting/Details}
    \item[] Question: Does the paper specify all the training and test details (e.g., data splits, hyperparameters, how they were chosen, type of optimizer, etc.) necessary to understand the results?
    \item[] Answer: \answerYes{} % Replace by \answerYes{}, \answerNo{}, or \answerNA{}.
    \item[] Justification: The paper outlines the datasets, models, evaluation techniques, and hyperparameters, allowing a clear understanding of the experimental results. Details on the dataset splits, language families, and claim distribution are provided in Section 3 and Appendix A.1.

\item {\bf Experiment Statistical Significance}
    \item[] Question: Does the paper report error bars suitably and correctly defined or other appropriate information about the statistical significance of the experiments?
    \item[] Answer: \answerNo{} % Replace by \answerYes{}, \answerNo{}, or \answerNA{}.
    \item[] Justification: The paper does not report error bars or confidence intervals. While the results are based on accuracy measurements, the paper does not include formal statistical significance tests.

\item {\bf Experiments Compute Resources}
    \item[] Question: For each experiment, does the paper provide sufficient information on the computer resources (type of compute workers, memory, time of execution) needed to reproduce the experiments?
    \item[] Answer: \answerYes{} % Replace by \answerYes{}, \answerNo{}, or \answerNA{}.
    \item[] Justification: The paper specifies the compute resources used, including MacBook Pros with Apple M1 Pro/Max and Intel Core i9 processors for inference, and a GPU cluster with NVIDIA A100 GPUs for pre-training mBERT, along with estimates of compute time (150 GPU-hours for pre-training and 100 CPU-hours for inference). Please see Appendix A.5.
    
\item {\bf Code Of Ethics}
    \item[] Question: Does the research conducted in the paper conform, in every respect, with the NeurIPS Code of Ethics \url{https://neurips.cc/public/EthicsGuidelines}?
    \item[] Answer: \answerYes{} % Replace by \answerYes{}, \answerNo{}, or \answerNA{}.
    \item[] Justification: The paper conforms to the NeurIPS Code of Ethics. It addresses the impact of translation bias on low-resource languages and emphasizes the importance of fairer language technologies. The paper’s conclusions are in line with ethical guidelines.

\item {\bf Broader Impacts}
    \item[] Question: Does the paper discuss both potential positive societal impacts and negative societal impacts of the work performed?
    \item[] Answer: \answerYes{} % Replace by \answerYes{}, \answerNo{}, or \answerNA{}.
    \item[] Justification: The paper highlights the positive societal impact of reducing bias in multilingual LLMs, which can improve the accessibility of accurate information. It also discusses the risks of spreading misinformation in low-resource languages. See Ethics Statement.
    
\item {\bf Safeguards}
    \item[] Question: Does the paper describe safeguards that have been put in place for responsible release of data or models that have a high risk for misuse (e.g., pre-trained language models, image generators, or scraped datasets)?
    \item[] Answer: \answerNA{} % Replace by \answerYes{}, \answerNo{}, or \answerNA{}.
    \item[] Justification: The paper does not involve the release of potentially harmful models or data, and thus this question does not apply.

\item {\bf Licenses for existing assets}
    \item[] Question: Are the creators or original owners of assets (e.g., code, data, models), used in the paper, properly credited and are the license and terms of use explicitly mentioned and properly respected?
    \item[] Answer: \answerYes{} % Replace by \answerYes{}, \answerNo{}, or \answerNA{}.
    \item[] Justification: The paper references all datasets and models used, such as the X-Fact dataset and pre-existing LLMs (e.g., GPT-4, mBERT), with proper citations. See Section 3.

\item {\bf New Assets}
    \item[] Question: Are new assets introduced in the paper well documented and is the documentation provided alongside the assets?
    \item[] Answer: \answerNA{} % Replace by \answerYes{}, \answerNo{}, or \answerNA{}.
    \item[] Justification: The paper does not introduce new assets.

\item {\bf Crowdsourcing and Research with Human Subjects}
    \item[] Question: For crowdsourcing experiments and research with human subjects, does the paper include the full text of instructions given to participants and screenshots, if applicable, as well as details about compensation (if any)? 
    \item[] Answer: \answerNA{} % Replace by \answerYes{}, \answerNo{}, or \answerNA{}.
    \item[] Justification: The paper does not involve crowdsourcing or human subjects.

\item {\bf Institutional Review Board (IRB) Approvals or Equivalent for Research with Human Subjects}
    \item[] Question: Does the paper describe potential risks incurred by study participants, whether such risks were disclosed to the subjects, and whether Institutional Review Board (IRB) approvals (or an equivalent approval/review based on the requirements of your country or institution) were obtained?
    \item[] Answer: \answerNA{} % Replace by \answerYes{}, \answerNo{}, or \answerNA{}.
    \item[] Justification: The paper does not involve research with human subjects.

\end{enumerate}

\end{document}